\documentclass[11pt,a4paper]{article}
\usepackage[hyperref]{emnlp2018}
\usepackage{times}
\usepackage{latexsym}
\usepackage{amsmath}
\usepackage{amssymb}
\usepackage{graphicx}
\usepackage{url}
\usepackage{booktabs}
\usepackage{multirow}
\usepackage{subfig} 
\usepackage[disable]{todonotes}
\usepackage{verbatim}

\usepackage{array}
\newcolumntype{L}[1]{>{\raggedright\let\newline\\\arraybackslash\hspace{0pt}}m{#1}}
\newcolumntype{C}[1]{>{\centering\let\newline\\\arraybackslash\hspace{0pt}}m{#1}}
\newcolumntype{R}[1]{>{\raggedleft\let\newline\\\arraybackslash\hspace{0pt}}m{#1}}

\newcommand{\squishlist}{
 \begin{list}{$\bullet$}
  { \setlength{\itemsep}{0pt}
     \setlength{\parsep}{3pt}
     \setlength{\topsep}{3pt}
     \setlength{\partopsep}{0pt}
     \setlength{\leftmargin}{1.5em}
     \setlength{\labelwidth}{1em}
     \setlength{\labelsep}{0.5em} } }

\newcommand{\squishend}{
  \end{list}  }

\aclfinalcopy 

\title{DeClarE: Debunking Fake News and False Claims \\using Evidence-Aware Deep Learning}

\author{Kashyap Popat$^1$, Subhabrata Mukherjee$^2$, Andrew Yates$^1$, Gerhard Weikum$^1$ \\
	$^{1}$Max Planck Institute for Informatics, Saarbr{\"u}cken, Germany \\
	$^{2}$Amazon Inc., Seattle, USA \\
	{\tt \{kpopat,ayates,weikum\}@mpi-inf.mpg.de, subhomj@amazon.com}
}

\date{}

\begin{document}
\maketitle
\begin{abstract}
Misinformation such as fake news is one of the big challenges of our society.
Research on automated fact-checking has proposed methods based on supervised learning, but these approaches do not consider external evidence apart from labeled training instances. Recent approaches counter this deficit by
considering external sources related to a claim. However, these methods require
substantial feature modeling and rich lexicons. This paper overcomes these limitations of prior work with an end-to-end model for evidence-aware credibility assessment of arbitrary textual claims, without any human intervention.
It presents a neural network model that judiciously aggregates signals from 
external evidence articles, the language of these articles and the trustworthiness of their sources. It also derives informative features for generating
user-comprehensible explanations that makes the neural network predictions transparent to the end-user. Experiments with four datasets and ablation studies
show the strength of our method.%
\end{abstract}

\section{Introduction}
{\bf Motivation:}
Modern media (e.g., news feeds, microblogs, etc.)
exhibit an increasing fraction of misleading 
and 
manipulative content, from 
questionable claims and ``alternative facts'' to completely faked news.
The media landscape is becoming
a twilight zone and battleground.
This societal challenge has led to the rise of fact-checking and debunking websites, such as {\it Snopes.com} and {\it PolitiFact.com}, 
where people research claims, manually assess their credibility, and present their verdict along with evidence (e.g., background articles, quotations, etc.). However, this manual verification is 
time-consuming. 
To keep up with the scale and speed at which misinformation spreads, 
we need 
tools to automate this debunking process.

\noindent{\bf State of the Art and Limitations:}
Prior work on ``truth discovery'' 
(see \citet{Li:2016:STD:2897350.2897352} for survey)\footnote{As fully objective and unarguable truth is often elusive or ill-defined, 
we use the term {\em credibility} rather than ``truth''.} largely focused on structured facts, typically in the form of subject-predicate-object triples,
or on social media platforms like Twitter, Sina Weibo, etc. 
Recently, 
methods have been proposed to assess the credibility of claims in natural language form
\cite{Popat:2017:TLE:3041021.3055133,Truth_varying_shades,DBLP:conf/acl/Wang17},
such as news 
headlines, quotes from speeches,
blog posts, etc.

The methods geared for general text input address the problem in different ways. 
On the one hand, methods like \citet{Truth_varying_shades,DBLP:conf/acl/Wang17} train neural networks on labeled claims from sites like {\it PolitiFact.com},
providing credibility assessments without any explicit feature modeling. However, they use only the text of questionable claims and no external evidence or interactions that provide limited context for credibility analysis.
These approaches also do not offer any explanation of their verdicts.
On the other hand, \citet{Popat:2017:TLE:3041021.3055133} considers external evidence in the form of other articles
(retrieved from the Web) that confirm or refute a claim, and jointly assesses the language style (using subjectivity lexicons),
the trustworthiness of the sources,
and the credibility of the claim.
This is achieved via a pipeline of supervised classifiers.
On the upside, this method generates user-interpretable explanations by pointing to informative snippets of evidence articles.
On the downside, it requires substantial feature modeling and rich lexicons to detect bias and subjectivity in the language style.


\noindent{\bf Approach and Contribution:}
To overcome the limitations of the prior works, 
we present {\em DeClarE}\footnote{Debunking Claims with Interpretable Evidence},
an end-to-end neural network model
for assessing and explaining the credibility of arbitrary claims in natural-language text form.
Our approach combines the best of both families of prior methods.
Similar to  \citet{Popat:2017:TLE:3041021.3055133}, DeClarE incorporates external evidence or counter-evidence from the Web 
as well as signals from the language style and the trustworthiness of the underlying sources.
However, our method does not require any feature engineering, lexicons, or other manual intervention. \citet{Truth_varying_shades,DBLP:conf/acl/Wang17} also develop an end-to-end model, but DeClarE goes far beyond in terms of considering external evidence and joint interactions between several factors, 
and also in its ability to generate user-interpretable explanations in addition to highly accurate assessments. 
For example, given the natural-language input claim {\em ``the gun epidemic is the leading cause of death of young African-American men, more than the next nine causes put together''}
by Hillary Clinton, DeClarE draws on evidence from the Web to arrive at its verdict 
{\em credible},
and returns annotated snippets like the one in Table~\ref{tab:evidence} as explanation.
These snippets, which contain evidence in the form of statistics and assertions, are automatically extracted from web articles from sources of varying credibility.

Given an input claim, DeClarE searches for web articles related to the claim.
It considers the {\em context} of the claim via word embeddings and the (language of) {web articles}
captured via a bidirectional LSTM (biLSTM), while using an {\em attention} mechanism to focus on parts of the articles 
according to their relevance to the claim.
DeClarE then aggregates 
all the information about claim source, web article contexts, attention weights, and trustworthiness of the underlying sources 
to assess the claim. 
It also derives informative features for interpretability,
like source embeddings that capture trustworthiness
and salient words
captured via attention.
Key contributions of this paper are:
\squishlist
	\item {\bf Model:} An end-to-end neural network model which automatically assesses the credibility of 
natural-language claims,
without any hand-crafted features or lexicons.
	\item {\bf Interpretability:} An {\em attention} mechanism in our model that 
generates user-comprehensible explanations, making credibility verdicts transparent and interpretable.
	\item {\bf Experiments:} Extensive experiments on four datasets and ablation studies, demonstrating  
effectiveness of our method over state-of-the-art baselines. 
\squishend

\begin{figure*}[t]
	\centering
	\includegraphics[width=0.78 \textwidth]{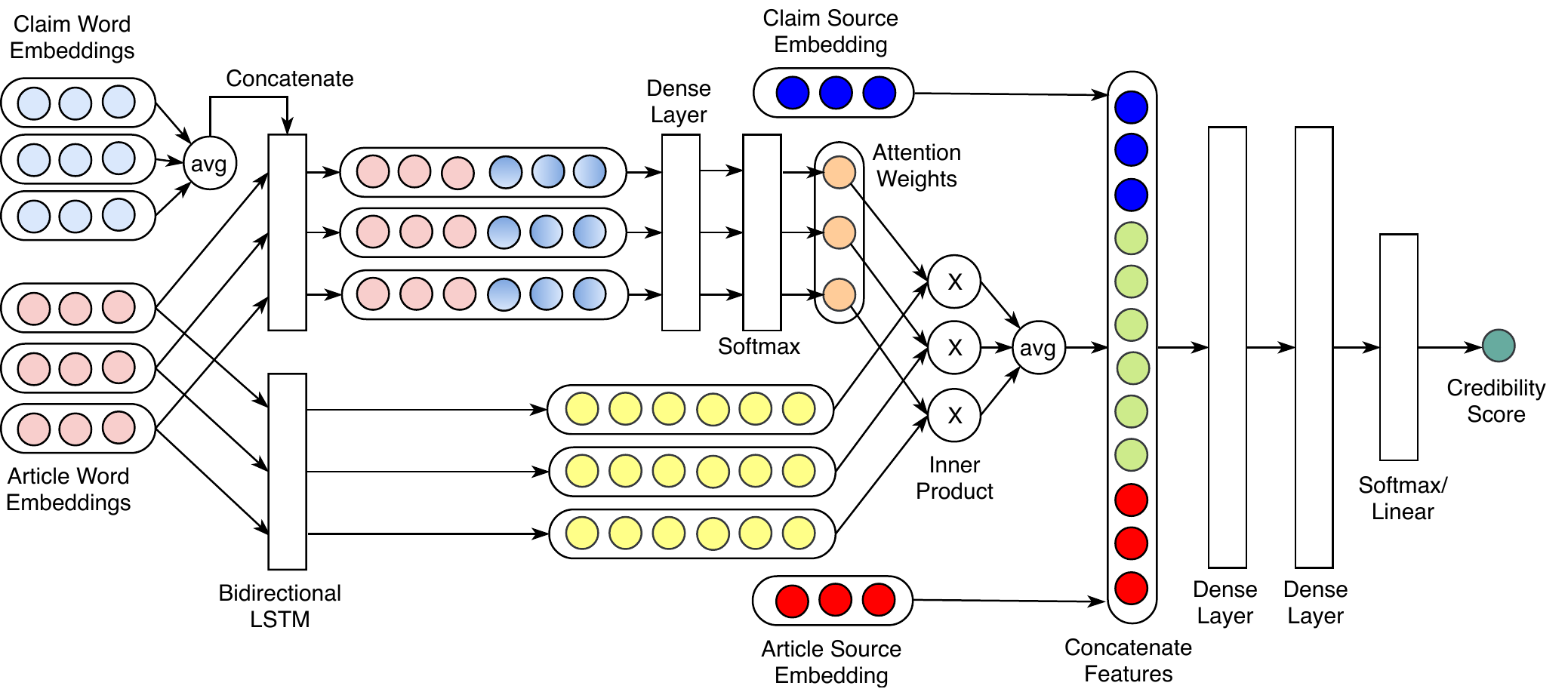}
	\vspace{1.5em}
	\caption{Framework for credibility assessment. Upper part of the pipeline combines the article and claim embeddings to get the claim specific attention weights. Lower part of the pipeline captures the article representation through biLSTM. Attention focused article representation along with the source embeddings are passed through dense layers to predict the credibility score of the claim.}
	\label{fig:arch}
\end{figure*}


\section{End-to-end Framework for Credibility Analysis}

Consider a set of $N$ claims $\langle C_n\rangle$ from the respective origins/sources $\langle CS_n\rangle$, where $n\in [1,N]$. Each claim $C_n$ is reported by a set of $M$ articles $\langle A_{m,n} \rangle$ along with their respective sources $\langle AS_{m,n} \rangle$, where $m \in [1,M]$. Each  corresponding tuple of claim and its origin, reporting articles and article sources -- $\langle C_n, CS_n, A_{m,n}, AS_{m,n} \rangle$ forms a training instance in our setting, along with the credibility label of the claim used as ground-truth during network training. Figure~\ref{fig:arch} gives a pictorial overview of our model. 
In the following sections, we provide a detailed description of our approach.

\subsection{Input Representations}
The input claim $C_n$ of length $l$ is represented as $[c_1, c_2,..., c_l]$ where $c_l \in \Re^d$ is the $d$-dimensional word embedding of the $l$-th word in the input claim. The source/origin of the claim $CS_n$ is represented by a $d_s$-dimensional embedding vector $cs_{n}\in \Re^{d_s}$. 

A reporting article $A_{m,n}$ consisting of $k$ tokens is represented by $[a_{m,n,1}, a_{m,n,2},..., a_{m,n,k}]$, where $a_{m,n,k} \in \Re^d$ is the $d$-dimensional word embedding vector for the $k$-th word in the reporting article $A_{m,n}$. The claim and article word embeddings have shared parameters. 
The source of the reporting article $AS_{m,n}$ is represented as a $d_s$-dimensional vector, $as_{m,n} \in \Re^{d_s}$. For the sake of brevity, we drop the notation subscripts $n$ and $m$ in the following sections by considering only a single training instance -- the input claim $C_n$ from source $CS_n$, the corresponding article $A_{m,n}$ and its sources $AS_{m,n} $ given by: $\langle C, CS, A, AS \rangle$.

\subsection{Article Representation}
To create a representation of an article, which may capture task-specific features such as whether it contains objective language,
we use a bidirectional Long Short-Term Memory (LSTM) network as proposed by \citet{Graves:2005:BLN:1986079.1986220}.  A basic LSTM cell consists of various gates to control the flow of information through timesteps in a sequence, making LSTMs suitable
for capturing long and short range dependencies in text that may be difficult to capture with standard recurrent neural networks (RNNs). Given an input word embedding of tokens $\langle a_k \rangle$, an LSTM cell performs various non-linear transformations to generate a hidden vector state $h_k$ for each token at each timestep $k$.

We use bidirectional LSTMs in place of standard LSTMs. Bidirectional LSTMs capture both the previous timesteps (past features) and the future timesteps (future features) via forward and backward states respectively. Correspondingly, there are two hidden states that capture past and future information that are concatenated to form the final output as:
$h_k =  [\overrightarrow{h_k}, \overleftarrow{h_k}]$.

\subsection{Claim Specific Attention}
As we previously discussed, it is important to consider the relevance of an article with respect to the claim; specifically, focusing or {\em attending} to parts of the article that discuss the claim. This is in contrast to prior works~\cite{Popat:2017:TLE:3041021.3055133,Truth_varying_shades,DBLP:conf/acl/Wang17} that ignore either the article or the claim, and therefore miss out on this important interaction.

We propose an attention mechanism to help our model focus on salient words in the article with respect to the claim. To this end, we compute the importance of each term in an article with respect to an overall representation of the corresponding claim. Additionally, incorporating attention helps in making our model transparent and interpretable, because it provides a way to generate the most salient words in an article as evidence of our model's verdict.

Following \citet{DBLP:journals/corr/WietingBGL15a}, the
overall representation of an input claim is generated by taking an average of the word embeddings of all the words therein:
\begin{align*} 
\bar{c} &= \frac{1}{l}\sum_{l} c_l
\end{align*}

We combine this overall representation of the claim with each article term:
\begin{align*} 
\hat{a}_{k} &= a_{k} \oplus \bar{c}
\end{align*}
\noindent where, $\hat{a}_{k} \in \Re^{d+d}$ and $\oplus$ denotes the concatenate operation.
We then perform a transformation to obtain claim-specific representations of each article term:
\begin{align*} 
a'_{k} &= \mathbf{f} (W_a \hat{a}_k + b_a)
\end{align*}

\noindent where $W_a$ and $b_a$ are the corresponding weight matrix and bias terms, and $\mathbf{f}$ is an activation function\footnote{In our model, the \textit{tanh} activation function gives best results.}, such as $ReLU$, $tanh$, or the identity function.
Following this, we use a softmax activation to calculate an attention score $\alpha_k$ for each word in the article capturing its relevance to the claim context:
\begin{align} 
\alpha_{k} &= \frac{\exp(a'_k)}{\sum_k \exp(a'_k)}
\label{eq:attn_weights}
\end{align}

\subsection{Per-Article Credibility Score of Claim}
Now that we have article term representations given by $\langle h_k \rangle$ and their relevance to the claim given by $\langle \alpha_k \rangle$, we need to combine them to predict the claim's credibility.
In order to create an attention-focused representation of the article considering both the claim and the article's language,
we calculate a weighted average of the hidden state representations for all article tokens based on their corresponding attention scores:
\begin{align} 
g &= \frac{1}{k} \sum_k{\alpha_k \cdot h_k}
\label{eq:art_rep}
\end{align}

We then combine all the different feature representations: the claim source embedding $(cs)$, the attention-focused article representation ($g$), and the article source embedding ($as$). In order to merge the different representations and capture their joint interactions, we process them with two fully connected layers with non-linear activations.
\begin{align} 
d_1 &= relu(W_c (g \oplus cs \oplus as) + b_c) \nonumber\\
d_2 & = relu(W_d d_1 + b_d) \nonumber
\end{align}

\noindent where, $W$ and $b$ are the corresponding weight matrix and bias terms. 

Finally, to generate the overall credibility label of the article for classification tasks, or credibility score for regression tasks, we process the final representation with a final fully connected layer:
\begin{align} 
\text{Classification: }\ s &= sigmoid(d_2) \\
\text{Regression: }\ s &= linear(d_2) 
\end{align}

\subsection{Credibility Aggregation}
The credibility score in the above step is obtained considering a single reporting article. As previously discussed, we have $M$ reporting articles per claim. Therefore, once we have the per-article credibility scores from our model, we take an average of these scores to generate the overall credibility score for the claim:
\begin{align} 
cred(C) &= \frac{1}{M} \sum_m{s_m}
\end{align}

\noindent This aggregation is done after the model is trained. 
\section{Datasets}
\label{sec:data}
We evaluate our approach and demonstrate its generality by performing experiments on four different datasets: a general fact-checking website, a political fact-checking website, a news review community, and a SemEval Twitter rumour dataset.

\subsection{Snopes} 
Snopes (\url{www.snopes.com}) is a general fact-checking website where editors manually investigate various kinds of rumors reported on the Internet. We used the Snopes dataset provided by \citet{Popat:2017:TLE:3041021.3055133}. This dataset consists of rumors analyzed on the Snopes website along with their credibility labels (\textit{true} or \textit{false}), sets of reporting articles, and their respective web sources. 

\subsection{PolitiFact} 	 
PolitiFact is a political fact-checking website (\url{www.politifact.com}) in which editors rate the credibility of claims made by various political figures in US politics. We extract all articles from PolitiFact published before December 2017. Each article includes a claim, the speaker (political figure) who made the claim, and the claim's credibility rating provided by the editors. 

PolitiFact assigns each claim to one of six possible ratings: \textit{true, mostly true, half true, mostly false, false} and \textit{pants-on-fire}. Following \citet{Truth_varying_shades}, we combine \textit{true, mostly true} and \textit{half true} ratings into the class label \textit{true} and the rest as \textit{false} -- hence considering only binary credibility labels. To retrieve the reporting articles for each claim (similar to \citet{Popat:2017:TLE:3041021.3055133}), we issue each claim as a query to a search engine\footnote{We use the Bing search API.} and retrieve the top 30 search results with their respective web sources. 

\subsection{NewsTrust}
NewsTrust is a news review community in which members review the credibility of news articles. We use the NewsTrust dataset made available by \citet{Mukherjee:2015:LJI:2806416.2806537}. This dataset contains NewsTrust stories from May 2006 to May 2014. Each story consists of a news article along with its source, and a set of reviews and ratings by community members. NewsTrust aggregates these ratings and assigns an overall credibility score (on a scale of 1 to 5) to the posted article. 
We map the attributes in this data to the inputs expected by DeClarE as follows: the title and the web source of the posted (news) article are mapped to the input claim and claim source, respectively. Reviews and their corresponding user identities are mapped to reporting articles and article sources, respectively. We use this dataset for the regression task of predicting the credibility score of the posted article.

\subsection{SemEval-2017 Task 8}
As the fourth dataset, we consider the benchmark dataset released by SemEval-2017 for the task of determining credibility and stance of social media content (Twitter) \cite{DBLP:conf/semeval/DerczynskiBLPHZ17}. The objective of this task is to predict the credibility of a questionable tweet (\textit{true}, \textit{false} or \textit{unverified}) along with a confidence score from the model. It has two sub-tasks: (i) a \textit{closed} variant in which models only consider the questionable tweet, and (ii) an \textit{open} variant in which models consider both the questionable tweet and additional context consisting of snapshots of relevant sources retrieved immediately before the rumor was reported, a snapshot of an associated Wikipedia article, news articles from digital news outlets, and preceding tweets about the same event. 
Testing and development datasets provided by organizers have 28 tweets (1021 reply tweets) and 25 tweets (256 reply tweets), respectively.

\begin{table}[t]
	\centering
	\small
	\resizebox{\columnwidth}{!}{
		\begin{tabular}{L{2.6cm}C{0.8cm}C{0.8cm}C{0.8cm}C{0.7cm}}
			\toprule
			\textbf{Dataset} & \textbf{SN} & \textbf{PF} & \textbf{NT} & \textbf{SE}\\
			\midrule
			Total claims & 4341 & 3568 & 5344 & 272 \\
			\hspace{1em}True claims & 1164 & 1867 & - & 127 \\
			\hspace{1em}False claims & 3177 & 1701 &  - & 50\\
			\hspace{1em}Unverified claims & - & - & - & 95 \\
			\midrule
			Claim sources & - & 95 & 161 & 10\\
			\midrule
			Articles & 29242 & 29556 & 25128 & 3717\\
			Article sources & 336 & 336 & 251 & 89\\
			\bottomrule
		\end{tabular}
	}
	\caption{Data statistics (SN: Snopes, PF: PolitiFact, NT: NewsTrust, SE: SemEval).}\label{tab:data_stats}
\end{table}

\subsection{Data Processing}
In order to have a minimum support for training, claim sources with less than 5 claims in the dataset are grouped into a single dummy claim source, and article sources with less than 10 articles are grouped similarly (5 articles for SemEval as it is a smaller dataset).

For Snopes and PolitiFact, we need to extract relevant snippets from the reporting articles for a claim. Therefore, we extract snippets of 100 words from each reporting article having the  maximum relevance score: $sim = sim_\text{bow} \times sim_\text{semantic}$

\noindent where $sim_\text{bow}$ is the fraction of claim words that are present in the snippet, and $sim_\text{semantic}$ represents the cosine similarity between the average of claim word embeddings and snippet word embeddings. We also enforce a constraint that the $sim$ score is at least $\delta$. 
We varied $\delta$ from 0.2 to 0.8 and found 0.5 to give the optimal performance on a withheld dataset. 
We discard all articles related to Snopes and PolitiFact websites from our datasets to have an unbiased model. 
Statistics of the datasets after pre-processing is provided in Table~\ref{tab:data_stats}. All the datasets are made publicly available at {\url{https://www.mpi-inf.mpg.de/dl-cred-analysis/}}.

\section{Experiments}
\label{sec:experiments}
We evaluate our approach by conducting experiments on four datasets, as described in the previous section. We describe our experimental setup and report our results in the following sections. 

\begin{table}[t]
	\centering
	\resizebox{\columnwidth}{!}{
		\begin{tabular}{L{5.2cm}C{0.6cm}C{0.6cm}C{0.6cm}C{0.6cm}}
			\toprule
			\textbf{Parameter} & \textbf{SN} & \textbf{PF} & \textbf{NT} & \textbf{SE} \\
			\midrule
			Word embedding length & 100 & 100 & 300 & 100 \\
			Claim source embedding length & - & 4 & 8 & 4\\
			Article source embedding length & 8 & 4 & 8 & 4\\
			LSTM size (for each pass)& 64 & 64 & 64 & 16\\
			Size of fully connected layers & 32 & 32 & 64 & 8\\		
			Dropout & 0.5 & 0.5 & 0.3 & 0.3\\
			\bottomrule
		\end{tabular}
	}
	\caption{Model parameters used for each dataset (SN: Snopes, PF: PolitiFact, NT: NewsTrust, SE: SemEval).}	\label{tab:tune_params}
\end{table}

\begin{table*}[t]
	\centering
	\small
	\begin{tabular}{L{1.5cm}L{3.2cm}C{2cm}C{2cm}C{1.5cm}C{0.9cm}}
		\toprule
		\textbf{Dataset} & \textbf{Configuration} & \textbf{\textit{True} Claims Accuracy (\%)} & \textbf{\textit{False} Claims Accuracy (\%)}  & \textbf{Macro F1-Score} & \textbf{AUC} \\	
		\midrule
		\multirow{7}{*}{Snopes}	& LSTM-text & 64.65 & 64.21 & 0.66 & 0.70\\
		& CNN-text & 67.15 & 63.14 & 0.66 & 0.72 \\
		& Distant Supervision & {\bf 83.21} & {\bf 80.78} & {\bf 0.82} & {\bf 0.88} \\
		\cmidrule{2-6}
		& DeClarE (Plain) & 74.37 & 78.57 & 0.78 & 0.83 \\
		& DeClarE (Plain+Attn) & 78.34 & 78.91 & 0.79 & 0.85  \\
		& DeClarE (Plain+SrEmb) & 77.43  & 79.80 & 0.79 & 0.85 \\
		& DeClarE (Full) & 78.96 & 78.32 & 0.79 & 0.86 \\
		\midrule
		\multirow{7}{*}{PolitiFact} & LSTM-text & 63.19 & 61.96 & 0.63 & 0.66 \\
		& CNN-text & 63.67 & 63.31 & 0.64 & 0.67 \\
		& Distant Supervision & 62.53 & 62.08 & 0.62 & 0.68 \\
		\cmidrule{2-6}
		& DeClarE (Plain) & 62.67 & 69.05 & 0.66 & 0.70 \\
		& DeClarE (Plain+Attn) & 65.53 & 68.49 & 0.66 & 0.72 \\
		& DeClarE (Plain+SrEmb) & 66.71 & 69.28 & 0.67 & 0.74 \\
		& DeClarE (Full) & \textbf{67.32} & \textbf{69.62 }& \textbf{0.68} & \textbf{0.75} \\
		\bottomrule
	\end{tabular}
	\caption{Comparison of various approaches for credibility classification on Snopes and PolitiFact datasets. 
	} \label{tab:snopes+politifact_result}

\end{table*}
\subsection{Experimental Setup}
When using the Snopes, PolitiFact and NewsTrust datasets, we reserve 10\% of the data as validation data for parameter tuning. We report 10-fold cross validation results on the remaining 90\% of the data; the model is trained on 9-folds and the remaining fold is used as test data. When using the SemEval dataset, we use the data splits provided by the task's organizers. The objective for Snopes, PolitiFact and SemEval experiments is binary (credibility) classification, while for NewsTrust the objective is to predict the credibility score of the input claim on a scale of 1 to 5 (i.e., credibility regression).
We represent terms using pre-trained GloVe Wikipedia 6B word embeddings \cite{pennington2014glove}. 
Since our training datasets are not very large, we do not tune the word embeddings during training.
The remaining model parameters are tuned on the validation data; the parameters chosen are reported in Table~\ref{tab:tune_params}.
We use Keras with a Tensorflow backend to implement our system. All the models are trained using Adam optimizer \cite{DBLP:journals/corr/KingmaB14} (learning rate: 0.002) with categorical cross-entropy loss for classification and  mean squared error loss for regression task. We use L2-regularizers with the fully connected layers as well as dropout. For all the datasets, the model is trained using each claim-article pair as a separate training instance.

To evaluate and compare the performance of DeClarE with other state-of-the-art methods, we report the following measures: 
\squishlist
	\item Credibility Classification (Snopes, PolitiFact and SemEval): accuracy of the models in classifying \textit{true} and \textit{false} claims separately, macro F1-score and Area-Under-Curve (AUC) for the ROC (Receiver Operating Characteristic) curve.
	\item Credibility Regression (NewsTrust): Mean Square Error (MSE) between the predicted and true credibility scores.
\squishend

\subsection{Results: Snopes and Politifact}
\label{subsec:snopes}

We compare our approach with the following state-of-the-art models: (i) LSTM-text, a recent approach proposed by \citet{Truth_varying_shades}.
(ii)  CNN-text: a CNN based approach proposed by \citet{DBLP:conf/acl/Wang17}.
(iii) Distant Supervision: state-of-the-art distant supervision based approach proposed by \citet{Popat:2017:TLE:3041021.3055133}. (iv) DeClare (Plain): our approach with only  biLSTM (no attention and source embeddings). (v) DeClarE (Plain+Attn): our approach with only biLSTM and attention (no source embeddings). (vi) DeClarE (Plain+SrEmb): our approach with only biLSTM and source embeddings (no attention). (vii) DeClarE (Full): end-to-end system with biLSTM, attention and source embeddings.

The results when performing credibility classification on the Snopes and PolitiFact datasets are shown in 
Table~\ref{tab:snopes+politifact_result}. 
DeClarE outperforms LSTM-text and CNN-text models by a large margin on both datasets.
On the other hand, for the Snopes dataset, performance of DeClarE (Full) is slightly lower than the Distant Supervision configuration (p-value of 0.04 with a pairwise t-test). However, the advantage of DeClarE over Distant Supervision approach is that it does not rely on hand crafted features and lexicons, and can generalize well to arbitrary domains without requiring any seed vocabulary. It is also to be noted that both of these approaches use external 
evidence
in the form of reporting articles discussing the claim, which are not available to the LSTM-text and CNN-text baselines. This 
demonstrates the value of external 
evidence  
for credibility assessment.

On the PolitiFact dataset, DeClarE outperforms all the baseline models by a margin of 7-9\% AUC (p-value of $9.12\mathrm{e}{-05}$ with a pairwise  t-test) with similar improvements in terms of Macro F1. 
A performance comparison of DeClarE's various configurations indicates the contribution of each component of our model, i.e, biLSTM capturing article representations, attention mechanism and source embeddings. The additions of both the attention mechanism and source embeddings improve performance over the plain configuration in all cases when measured by Macro F1 or AUC.

\subsection{Results: NewsTrust}
When performing credibility regression on
the NewsTrust dataset, 
we evaluate the models in terms of mean squared error (MSE; lower is better) for credibility rating prediction. 
We use the first three models described in Section~\ref{subsec:snopes} as baselines. For CNN-text and LSTM-text, we add a linear fully connected layer as the final layer of the model to support regression. Additionally, we also consider the state-of-the-art CCRF+SVR model based on Continuous Conditional Random Field (CCRF) and Support Vector Regression (SVR) proposed by \citet{Mukherjee:2015:LJI:2806416.2806537}.
The results are shown in Table~\ref{tab:nt_result}. We observe that DeClarE (Full) outperforms all four baselines, with a 17\% decrease in MSE compared to the best-performing baselines (i.e., LSTM-text and Distant Supervision). The DeClarE (Plain) model performs substantially worse than the full model, illustrating the value of including attention and source embeddings.
CNN-text performs substantially worse than the other baselines.

\begin{table}[t]
	\centering
	\begin{tabular}{L{3cm}C{1.5cm}}
		\toprule
		\textbf{Configuration} & \textbf{MSE} \\
		\midrule
		CNN-text & 0.53 \\
		CCRF+SVR & 0.36  \\
		LSTM-text & 0.35 \\
		DistantSup & 0.35 \\		
		DeClarE (Plain) & 0.34 \\
		DeClarE (Full) & {\bf 0.29}\\
		\bottomrule
	\end{tabular}
	\caption{Comparison of various approaches for credibility regression on NewsTrust dataset.
	}	\label{tab:nt_result}
\end{table}

\begin{figure*}[t]
	\centering
	\subfloat[Projections of article representations using PCA; DeClarE obtains clear separation between representations of non-credible articles (\textit{red}) vs. true ones (\textit{green}).]{%
		\includegraphics[width=0.31\linewidth]{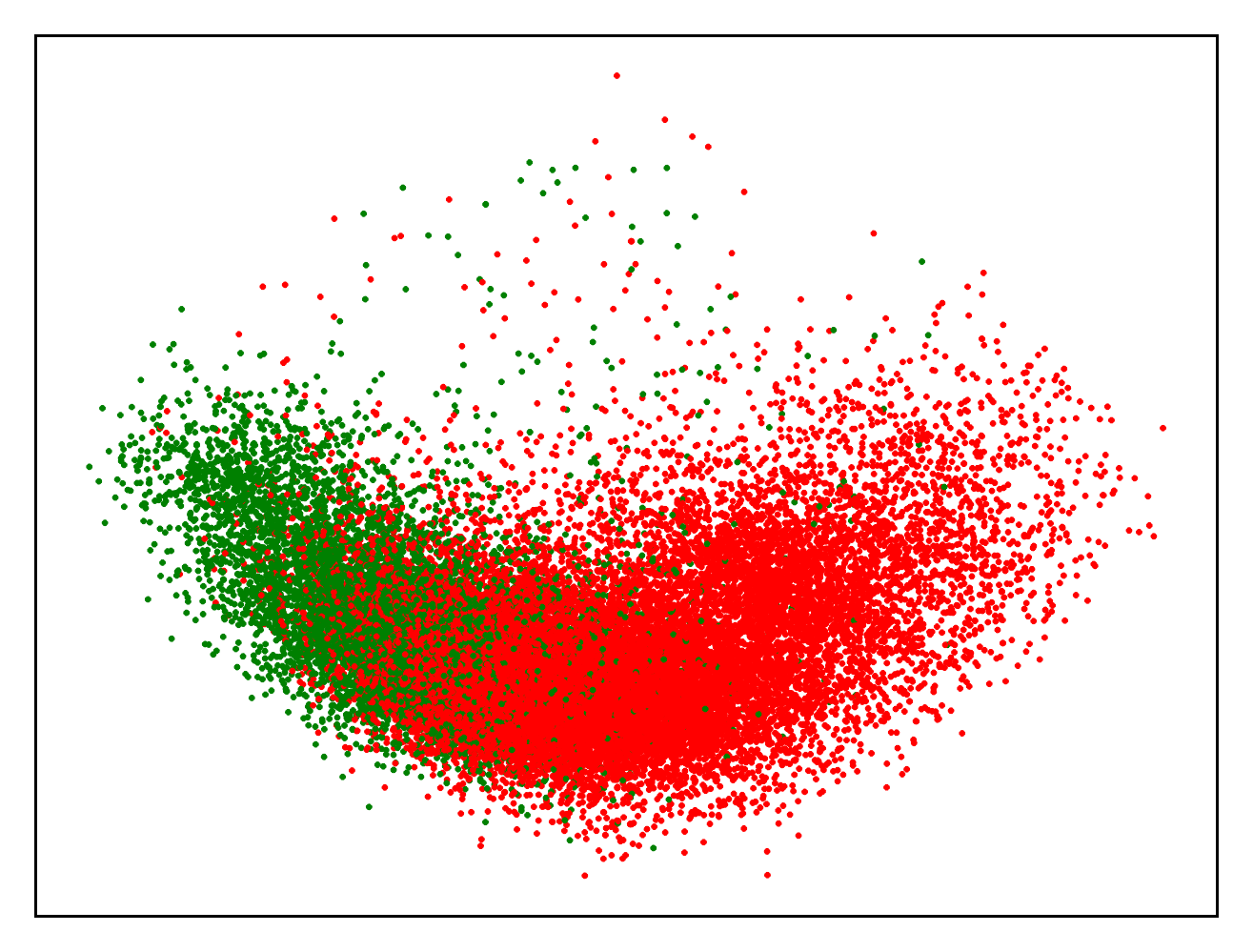}
		\label{fig:2a}
	}
	\hfill
	\subfloat[Projections of article source representations using PCA; DeClarE clearly separates fake news sources from authentic ones.
	]{%
		\includegraphics[width=0.31\linewidth]{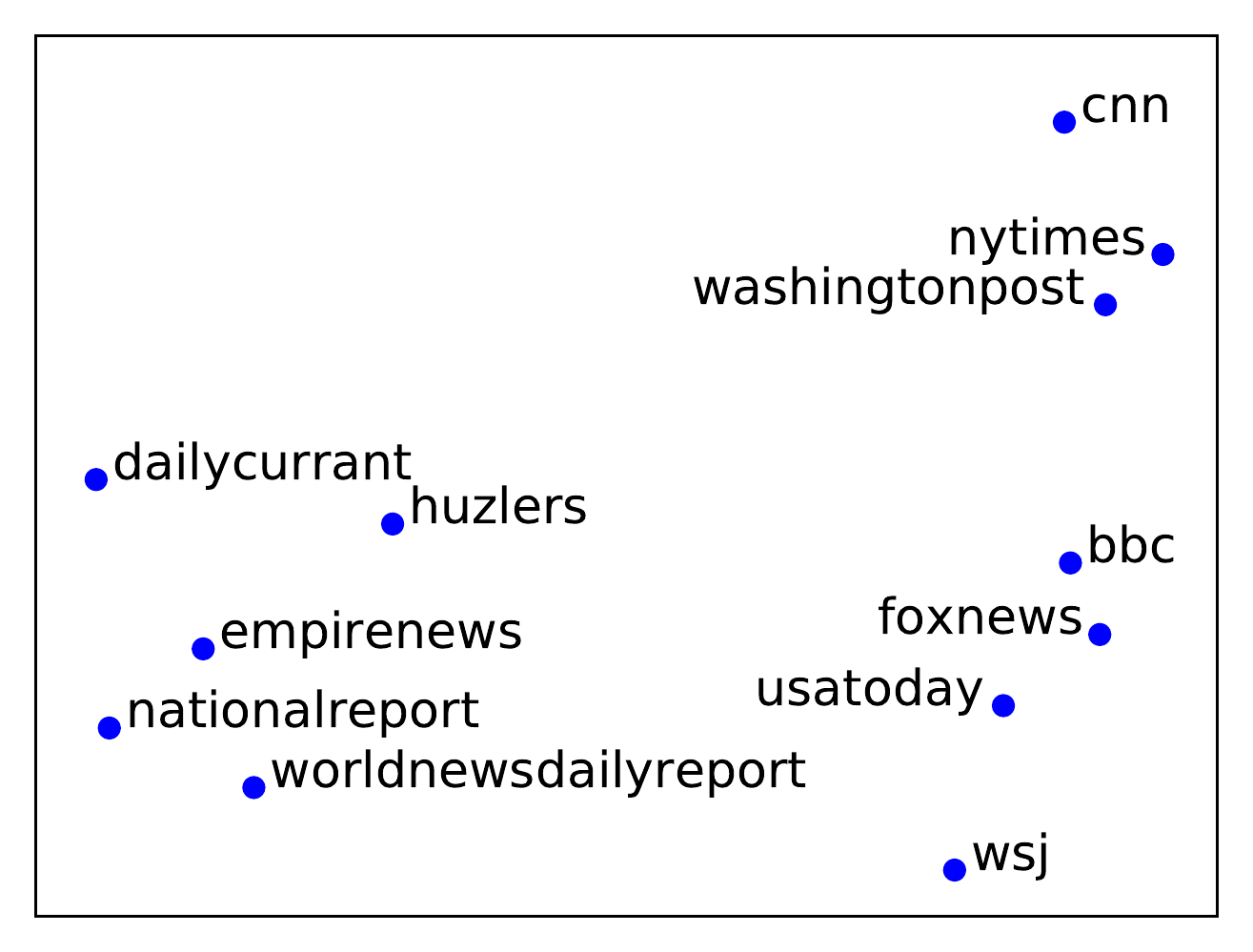}
		\label{fig:2b}
	}
	\hfill
	\subfloat[Projections of claim source representations using PCA; DeClarE clusters politicians of similar ideologies close to each other in the embedding space.
	]{%
		\includegraphics[width=0.31\linewidth]{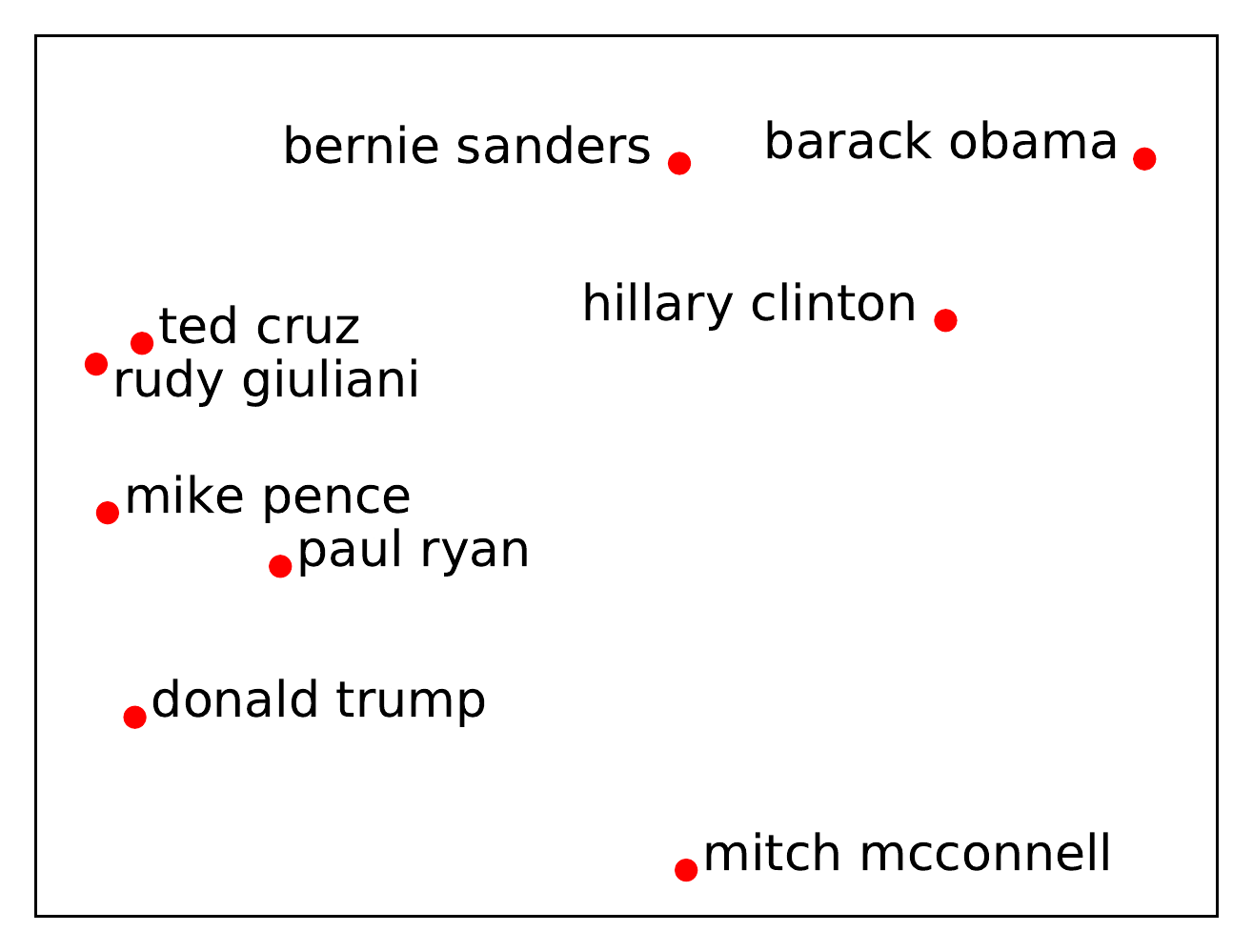}
		\label{fig:2c}
	}
	\caption{Dissecting the article, article source and claim source representations learned by DeClarE.}
	\label{fig:2}
\end{figure*}

\subsection{Results: SemEval}
On the SemEval dataset, the objective is to perform credibility classification of a tweet while also producing a classification confidence score.
We compare the following approaches and consider both variants of the SemEval task:
(i) \textit{NileTMRG} \cite{DBLP:conf/semeval/EnayetE17}: the best performing approach for the \textit{close} variant of the task, (ii) \textit{IITP} \cite{DBLP:conf/semeval/SinghNAEB17}: the best performing approach for the \textit{open} variant of the task, (iii) DeClare (Plain): our approach with only  biLSTM (no attention and source embeddings), and (iv) DeClarE (Full): our end-to-end system with biLSTM, attention and source embeddings.

We use the evaluation measure proposed by the task's organizers:
macro F1-score for overall classification and Root-Mean-Square Error (RMSE) over confidence scores.
Results are shown in Table~\ref{tab:se_result}.
We observe that DeClarE (Full) outperforms all the other approaches --- thereby, re-affirming its power in harnessing 
external evidence. 

\section{Discussion}
\subsection{Analyzing Article Representations}
In order to assess how our model separates articles reporting false claims from those reporting true ones, we employ dimensionality reduction using Principal Component Analysis (PCA) to project the article representations ($g$ in Equation~\ref{eq:art_rep}) from a high dimensional space to a 2d plane. The projections are shown in Figure~\ref{fig:2a}. We observe that DeClarE obtains clear separability between credible versus non-credible articles in Snopes dataset.

\begin{table}[t]
	\centering
	\begin{tabular}{L{3cm}C{1.7cm}C{1.2cm}}
		\toprule
		\textbf{Configuration} & \textbf{Macro Accuracy} & \textbf{RMSE} \\
		\midrule
		IITP (Open) & 0.39 & 0.746 \\		
		NileTMRG (Close) & {0.54} & 0.673\\
		DeClarE (Plain) & 0.46 & 0.687 \\	
		DeClarE (Full) &  {\bf 0.57} & {\bf 0.604} \\		
		\bottomrule
	\end{tabular}
	\caption{Comparison of various approaches for credibility classification on SemEval dataset. 
	}	\label{tab:se_result}
\end{table}

\subsection{Analyzing Source Embeddings}
Similar to the treatment of article representations, we perform an analysis with the claim and article source embeddings by employing PCA and plotting the projections. We sample a few popular news sources from Snopes and claim sources from PolitiFact. These news sources and claim sources are displayed in Figure~\ref{fig:2b} and Figure~\ref{fig:2c}, respectively. From Figure~\ref{fig:2b} we observe that DeClarE clearly separates fake news sources like \textit{nationalreport}, \textit{empirenews}, \textit{huzlers}, etc. from mainstream news sources like \textit{nytimes}, \textit{cnn}, \textit{wsj}, \textit{foxnews}, \textit{washingtonpost}, etc. Similarly, from Figure~\ref{fig:2c} we observe that DeClarE locates politicians with similar ideologies and opinions close to each other in the embedding space. 

\begin{table*}[t]
	\renewcommand{\arraystretch}{0.5}
		\begin{tabular}{C{0.96\linewidth}}
			\toprule
			\includegraphics[width=0.96\textwidth]{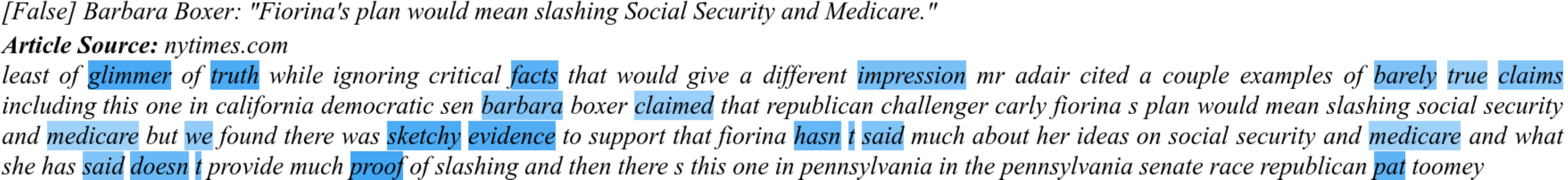}	\\
			\midrule
			\includegraphics[width=0.96\textwidth]{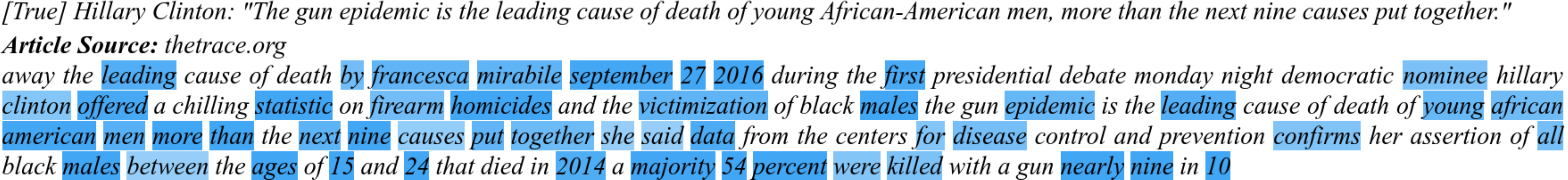}	\\
			\midrule
			\includegraphics[width=0.96\textwidth]{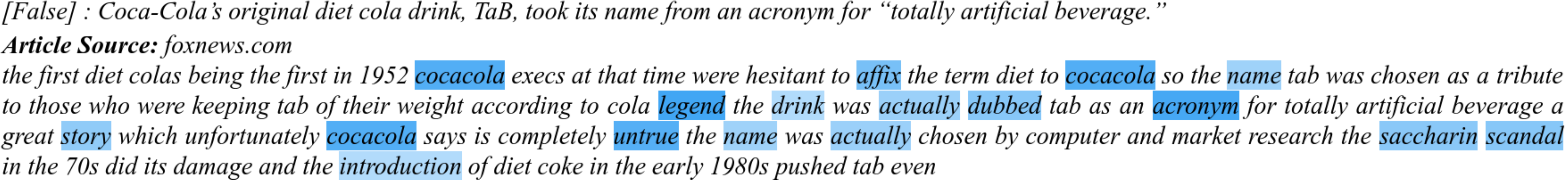}	\\
			\midrule
			\includegraphics[width=0.96\textwidth]{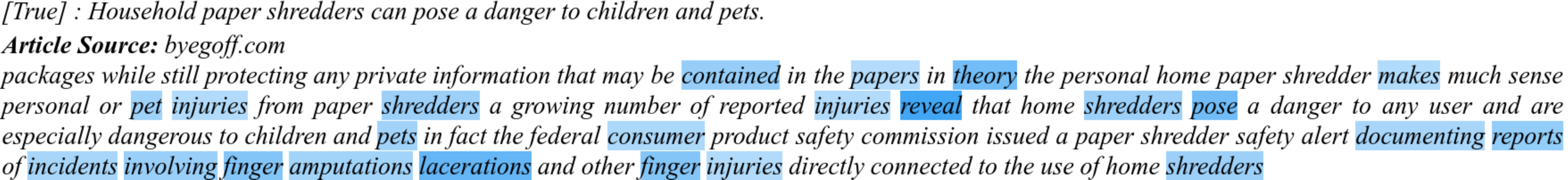}	\\
			\bottomrule
		\end{tabular}
	\caption{Interpretation via attention (weights) ([\textit{{True}}]/[\textit{{False}}] indicates the verdict from DeClarE).}
	\label{tab:evidence}
\end{table*}

\subsection{Analyzing Attention Weights}
Attention weights help understand what DeClarE focuses on during learning and how it affects its decisions -- thereby, making our model transparent to the end-users.
Table~\ref{tab:evidence} illustrates some interesting claims and salient words (highlighted) that DeClarE focused on during learning. Darker shades indicate higher weights given to the corresponding words. As illustrated in the table, DeClarE gives more attention to important words in the reporting article that are relevant to the claim and also play a major role in deciding the corresponding claim's credibility. 
In the first example on Table~\ref{tab:evidence}, highlighted words such as ``\textit{..barely true...}" and ``\textit{..sketchy evidence...}" help our system to identify the claim as \textit{not credible}. On the other hand, highlighted words in the last example, like, ``\textit{..reveal...}" and ``\textit{..documenting reports...}" help our system to assess the claim as \textit{credible}. 

\section{Related Work}
Our work is closely related to the following areas:

\noindent \textbf{Credibility analysis of Web claims:} 
Our work builds upon approaches for performing credibility analysis of natural language claims in an open-domain Web setting. The approach proposed in \citet{Popat:2016:CAT:2983323.2983661, Popat:2017:TLE:3041021.3055133} employs stylistic language features and the stance of articles to assess the credibility of the natural language claims. However, their model heavily relies on handcrafted language features. 
\citet{Truth_varying_shades, DBLP:conf/acl/Wang17} propose neural network based approaches for determining the credibility of a textual claim, but it does not consider external sources like web evidence and claim sources. These can be important evidence sources for credibility analysis. 
The method proposed by \citet{Samadi:2016:CIF:3015812.3015845} uses the Probabilistic Soft Logic (PSL) framework to estimate source reliability and claim correctness. \citet{Vydiswaran:2011:CTP:2020408.2020567} proposes an iterative algorithm which jointly learns the veracity of textual claims and trustworthiness of the sources. These approaches do not consider the deeper semantic aspects of language, however. 
\citet{Wiebe:2005:CSO:2132047.2132112, I11-1129, P13-1162} study the problem of detecting bias in language, but do not consider credibility.

\noindent \textbf{Truth discovery:} Prior approaches for truth discovery \cite{Yin:2008:TDM:1399100.1399392, Dong:2009:ICD:1687627.1687690, Dong:2015:KTE:2777598.2777603, Li:2011:TVT:2004686.2005589, Li:2014:CAT:2735496.2735505, Li:2015:DET:2783258.2783277,  DBLP:conf/ijcai/PasternackR11, Pasternack:2013:LCA:2488388.2488476, Ma:2015:FFG:2783258.2783314, Zhi:2015:MTE:2783258.2783339,DBLP:journals/pvldb/GaoLZFH15, Lyu:2017:TDC:3132847.3133069} have focused on structured data with the goal of addressing the problem of conflict resolution amongst multi-source data. 
\citet{DBLP:conf/acl/NakasholeM14} proposed a method to extract conflicting values from the Web in the form of Subject-Predicate-Object (SPO) triplets and uses language objectivity analysis to determine the true value. Like the other truth discovery approaches, however, this approach is mainly suitable for use with structured data.

\noindent \textbf{Credibility analysis in social media:} \citet{Mukherjee:2014:PDC:2623330.2623714, Mukherjee:2015:LJI:2806416.2806537} propose PGM based approaches to jointly infer a statement's credibility and the reliability of sources using language specific features. Approaches like \cite{Castillo:2011:ICT:1963405.1963500, Qazvinian:2011:RIM:2145432.2145602, Yang:2012:ADR:2350190.2350203,C12-2131, Gupta:2013:FSC:2487788.2488033, Zhao:2015:EME:2736277.2741637, P17-2102} propose supervised methods for detecting deceptive content in social media platforms like Twitter, Sina Weibo, etc. 
Similarly, approaches like \citet{Ma:2016:DRM:3061053.3061153, Ruchansky:2017:CHD:3132847.3132877} use neural network methods to identify fake news and rumors on social media. \citet{Kumar:2016:DWI:2872427.2883085} studies the problem of detecting hoax articles on Wikipedia. 
All these rely on domain-specific and community-specific features like retweets, likes, upvotes, etc.

\section{Conclusion}
In this work, we propose a completely automated end-to-end neural network model, DeClarE, for evidence-aware credibility assessment of natural language claims without requiring hand-crafted features or lexicons. DeClarE captures signals from external evidence articles and models joint interactions between various factors like the context of a claim, the language of reporting articles, and trustworthiness of their sources. Extensive experiments on real world datasets demonstrate our effectiveness over state-of-the-art baselines.

\bibliography{emnlp2018}
\bibliographystyle{acl_natbib_nourl}

\end{document}